\documentclass[final]{article}

\usepackage{nips_2017}

\usepackage[utf8]{inputenc} 
\usepackage[T1]{fontenc}    
\usepackage{hyperref}       
\usepackage{url}            
\usepackage{booktabs}       
\usepackage{amsfonts}       
\usepackage{nicefrac}       
\usepackage{microtype}      
\usepackage{amsmath,graphicx,siunitx,natbib, amsthm, amssymb}

\title{Interpretable Vector AutoRegressions with Exogenous Time Series}

\author{Ines Wilms\thanks{iw74@cornell.edu, ines.wilms@kuleuven.be, \url{http://feb.kuleuven.be/ines.wilms}} \\
	Department of Statistical Science \\
	Cornell University \\
	Faculty of Economics and Business \\
	KU Leuven
	\And
	Sumanta Basu\thanks{sumbose@cornell.edu, \url{http://faculty.bscb.cornell.edu/\~basu/}}\\
	Department of Statistical Science \\
	Cornell University \\
	\And
	Jacob Bien\thanks{jbien@usc.edu, \url{http://faculty.bscb.cornell.edu/\~bien/}} \\
	Data Sciences and Operations \\
	University of Southern California
	\And
	David S. Matteson\thanks{matteson@cornell.edu, \url{http://www.stat.cornell.edu/\~matteson/}} \\
	Department of Statistical Science \\
	Cornell University
}

\begin{document}
	\maketitle
	
	\begin{abstract}
		The Vector AutoRegressive (VAR) model is fundamental to the study of multivariate time series. 
		Although VAR models are intensively investigated by many researchers, 
		practitioners often show more interest in analyzing VARX models that incorporate the impact of unmodeled exogenous variables (X) into the VAR. 
		However, since the parameter space grows quadratically with the number of time series, estimation quickly becomes challenging.
		While several proposals have been made to sparsely estimate large VAR models, the estimation of large VARX models is under-explored. Moreover, typically these sparse proposals involve a lasso-type penalty and do not incorporate lag selection into the estimation procedure. 
		As a consequence, the resulting models may be difficult to interpret. 
		In this paper, we propose a lag-based hierarchically sparse estimator, called "HVARX", for large VARX models. 
		We illustrate the usefulness of HVARX on a cross-category management marketing application. Our results show how it
		 provides a highly interpretable model, and 
		 improves out-of-sample forecast accuracy compared to a lasso-type approach.
		
	\end{abstract}
	
	\section{Introduction}
	Autocorrelated multivariate time series are primarily modeled using Vector AutoRegressive (VAR) models.  Applications are found in diverse fields such as  machine learning (e.g., \citealp{Melnyk16}), macro-economics (e.g., \citealp{Banbura10}), and marketing (e.g., \citealp{Gelper15}).
	In a VAR$_k(p)$ model, a stationary $k$-dimensional vector
        time series $\{{y}_t\}$ is modeled in terms of its previous $p$
        values. In low-dimensional settings, in which $k^2p$ is small
        relative to the series length $T$, VAR models can be
        estimated using multivariate least squares. However, this approach becomes imprecise or even intractable in large VAR models, in
        which $k^2p$ can exceed $T$. 
Recent interest in modeling high dimensional time series has driven many advances in penalized estimators for large VARs  
	(e.g., \citealp{Davis15},  \citealp{Basu15},
        \citealp{Gelper15},  \citealp{Nicholson16}).  Most utilize 
        an $\ell_1$-penalty (lasso-penalty,
        \citealp{Tibshirani96}) to obtain sparse parameter estimates. 
	
	While VAR models are more intensively investigated,
        practitioners are often more interested in estimating "VARX"
        models, i.e., vector autoregressions with additional unmodeled exogenous (X) time series components. 
	In economics, for example, VARX models have become popular to model small open economies given U.S.\ variables. 
	The VARX is also of interest in scenarios where one desires forecasts from only a subset of a larger set of time series. In marketing, for instance, one may use the VARX to predict sales of several related products using their past sales, price and promotion information. 
	
	However, large VARX models - which are as heavily
        overparameterized as large VARs -  have received only limited
        attention in the literature (see, e.g., \citealt{Nicholson17}). 
	Moreover, endogenous and exogenous components have different lag orders, and lag selection by classical information criteria is intractable, in general. 
	While $\ell_1$-type estimators can resolve the dimensionality
        issue, they are not well suited to solve lag selection
        issues. The resulting models are not easily interpretable and
        \cite{Nicholson16} show that the prediction performance of these estimators substantially deteriorates for VARs with large lag orders as they select many high lag order coefficients. 
	
	To simultaneously address the dimensionality and lag selection
        issues while providing highly interpretable estimates, we propose a lag-based hierarchically sparse estimation procedure, "HVARX", for large VARX. 
	Like most papers on large VAR estimation, our parameter estimates are sparse to accommodate the dimensionality issue.
	Unlike other approaches, we embed automatic lag selection of
        both the exogenous and endogenous components into
        estimation. We build on the HVAR penalties of
        \cite{Nicholson16}, which use hierarchical group lasso
        penalties (e.g., \citealp{Yuan06} for group lasso) in place of
       $\ell_1$-penalties.  
	The proposed estimation procedure is very flexible: a separate
        lag order is learned for each endogenous and exogenous series in each marginal equation of the VARX.
The result is a parsimonious, highly
        interpretable model with low lag orders. 
	We apply HVARX to estimate cross-category demand effects among a large number of product categories for a particular U.S.\ grocery store. Our results show that HVARX  provides 
	highly interpretable results, and 
	considerably improves forecast accuracy compared to a lasso-type estimator.

	\section{Methodology}
	
	\paragraph{Model.}
	In a VARX$_{k, m}(p, s)$, a stationary $k$-dimensional vector series $\{{y}_t\}$ is modeled in terms of its own past $p$ values and the past $s$ values of a stationary $m$-dimensional vector series $\{{x}_t\}$: 
	\begin{equation}
	{y}_t =  \sum_{\ell=1}^p   \Phi_{\ell} {  y}_{t-\ell} + \sum_{j=1}^s   B_{j} {  x}_{t-j} +  {  \epsilon}_{t}, \label{VARX} 
	\end{equation} 
	where
	$ \{\Phi_{\ell} \in \mathbb{R}^{k \times k} \}_{\ell=1}^{p}$ are endogenous coefficient matrices,
	$ \{B_{j} \in \mathbb{R}^{k \times m} \}_{j=1}^{s}$ are  exogenous coefficient matrices, and
	$\{{\epsilon}_t\}$ is a $k$-dimensional mean-zero white noise  vector time series. 
	We assume, without loss of generality, that all time series
        are mean-centered and thus omit an intercept. 
	A VAR$_{k}(p)$ model is a special case of the VARX$_{k,m}(p,s)$ model in (\ref{VARX}) where $m=0$.
	In the remainder, we consider the VARX$_{k, m}(p, s)$ in compact matrix notation:
	${Y} =   {\Phi} {Z} + {B} {  X} +  {{E}},$ 
	where ${Y}=[{y}_{\bar{o}+1} \ldots {y}_T] \in \mathbb{R}^{k\times ({T-\bar{o}})},$ 
	${Z} = [{z}_{\bar o+1} \ldots {z}_T] \in \mathbb{R}^{kp\times ({T-\bar o})}$, with
	${z}_t = [{y}^\top_{t-1} \ldots {y}^\top_{t-{p}} ]^\top \in \mathbb{R}^{(kp\times 1)},$ 
	${X} = [{x}_{\bar{o}+1} \ldots {x}_T] \in \mathbb{R}^{ms\times ({T-\bar{o}})}$
	with 
	${x}_t = [ x^\top_{t-1} \ldots x^\top_{t-s} ]^\top \in \mathbb{R}^{(ms\times 1)}$,
	with $\bar{o}=\max(p, s)$, 
	${{E}} = [{\epsilon}_{\bar o+1}  \ldots {\epsilon}_T] \in \mathbb{R}^{k\times ({T-\bar o})}, 
	{\Phi} = [{\Phi}_{1}\ldots {  \Phi}_{p}] \in \mathbb{R}^{k\times kp},$ and
	${B} = [{B}_{1} \ldots {B}_{s}] \in \mathbb{R}^{k\times ms}.$ 
	
\begin{figure}
		\centering
		\includegraphics[width=8cm]{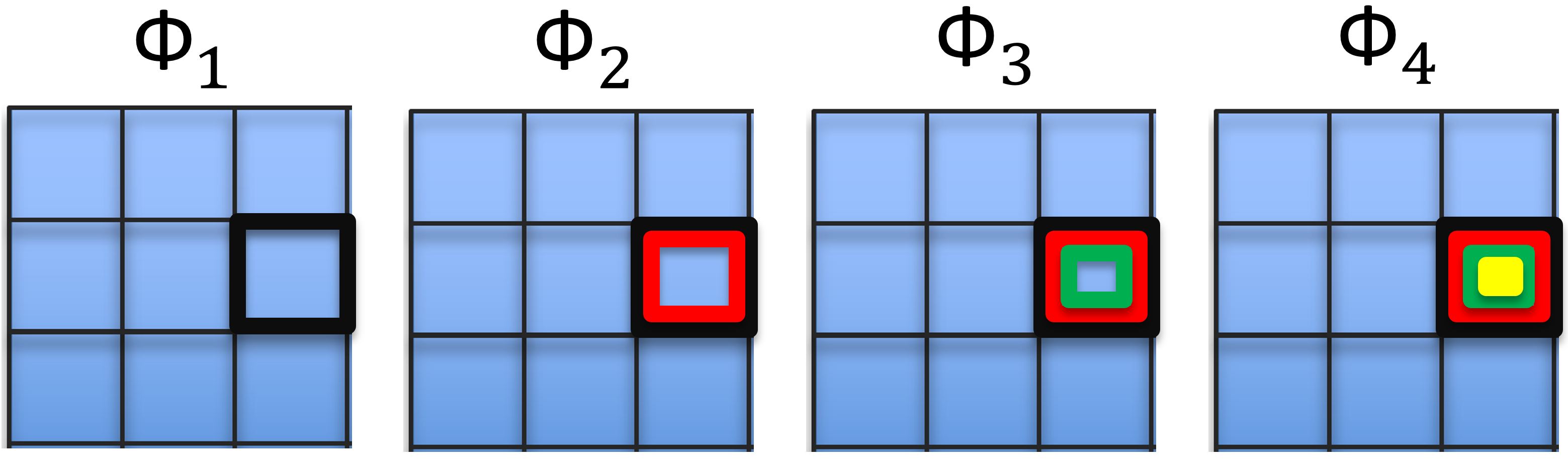}
		\caption{Lag-based hierarchical group structure of element (2,3) of the $\Phi$ matrices for an illustrative VARX model with $k=3$ and $p=4$. 
	The first group (black) contains element (2,3) of all $\Phi$ matrices. The second group (red) is nested in the first and only contains element (2,3) of ($\Phi_2, \Phi_3, \Phi_4$). 
	The nested structure continues with the green and yellow groups.\label{HLag}} 
	\vspace{-6pt}	
	\end{figure}

	\paragraph{Estimator.}
	To obtain estimates $(\widehat{\Phi}, \widehat{B})$, we use a
        sparsity inducing estimator which places a lag-based
        hierarchical group lasso (HLag; \citealt{Nicholson16}) penalty on both the endogenous and exogenous parameters. The parameter estimates are obtained as 
	\begin{equation}
	\small
	\!(\widehat{{\Phi}}, \widehat{{B}}) \!=\! \underset{{  \Phi},{  B}}{\operatorname{argmin}} \!\left \{\!\dfrac{1}{2} \|{Y} -  {\Phi} {Z} - {B}{X}\|_F^2  \!+\! \lambda_{{\Phi}} \sum_{i=1}^{k} \sum_{d=1}^{k} \sum_{\ell=1}^{p} \|{\Phi}_{(\ell:p), id} \|_2  \!+\! \lambda_{{B}} \sum_{i=1}^{k} \sum_{r=1}^{m} \sum_{j=1}^{s} \|{B}_{(j:s), ir} \|_2\!\right \}\!,
	\!\!\label{HVARXeq}
	\end{equation}
	where $\|\cdot\|_F$ denotes the Frobenius norm,  
	${\Phi}_{(\ell:p), id} = \left [{\Phi}_{\ell, id} \ldots  {\Phi}_{p, id} \right] \in \mathbb{R}^{(p-\ell+1)}$, and 
	${B}_{(j:s), ir} = \left [{B}_{j, ir} \ldots  {B}_{s, ir} \right] \in \mathbb{R}^{(s-j+1)}$. This defines the HVARX estimator.
	Figure \ref{HLag} illustrates how the nested group principle can be used to achieve hierarchical sparsity. 
	By using this hierarchical group structure, the estimator
        forces lower lagged coefficients of a particular time series
        to be selected before its higher order time lagged
        coefficients. As such, it automatically determines the maximum
        lag order for the corresponding component (for more, see
        \citealt{Nicholson16}).  HVARX is highly flexible as each (endogenous and exogenous) series has its own lag structure in each marginal equation of the VARX. 
	The penalty parameters $\lambda_{{\Pi}}>0$ and $\lambda_{{B}}>0$ regulate the degree of sparsity in $\widehat{{\Phi}}$ and  $\widehat{{B}}$, respectively,
 and are selected using a time-series cross-validation approach (see \citealt{Wilms17}).
	To compute HVARX, we use the Proximal Gradient Algorithm in \citeauthor{Wilms17} (\citeyear{Wilms17}, Algorithm 2). An implementation of our method is available in the \texttt{R} package \texttt{bigtime} \citep{bigtime}.
	
\paragraph{Theoretical Properties.}
Assuming the VARX process $\{y_t\}$ is stable with Gaussian innovations $\{\epsilon_t\}$, theoretical analysis of HVARX and "$\ell_1$-VARX", the standard lasso estimator of the VARX model, can be conducted under a suitable Restricted Eigenvalue (RE) assumption on the design matrix $\tilde{Z} = [Z^\top, X^\top]$ and a deviation bound on $\tilde{Z}^\top E$ (cf.\ \citealp{Basu15} for more details). The stability properties of a Gaussian VARX model have been studied recently in \citet{lin2017regularized}, along with the validity of necessary RE and deviation conditions for the $\ell_1$-VARX when the matrix $\Phi$ is sparse and the matrix $B$ is low-rank. A straightforward adaptation of these results, coupled with the deviation bounds in Proposition 2.4 of \citet{Basu15} shows that $\ell_1$-VARX can consistently estimate $\Phi$ and $B$ in a high-dimensional regime where the time series dimensions $k$ and $m$ grow exponentially with sample size $(T-\bar{o})$. Theoretical properties of HVARX in high-dimension can be explored along the same route as $\ell_1$-VARX, although additional challenges remain due to the non-separable nature of the HLag penalty. We expect to leverage recent advances in the theory of overlapping group Lasso to address these challenges.

	\section{Marketing Application}
	Retailers carry products in a large variety of categories such as soft drinks. 
	Successful cross-category management requires retailers to understand "cross-category demand effects", i.e., the effects of prices, promotions, and sales of one category on the sales (or demand) of another product category. 
	The VARX model can be used to estimate these demand effects,        with the sales series taken as the endogenous time series  and
        the price and promotion series taken as the exogenous time series. 
	
	\paragraph{Data.}
	We use data on 16 product categories\footnote{Beer, Bottled Juices, Refrigerated Juices, Frozen Juices, Soft Drinks, Crackers, Snack Crackers, Front end candies, Cookies, Cheeses, Canned Soup, Cereals, Oatmeal, Frozen Dinners, Frozen Entrees and Canned Tuna.} from a grocery store of Dominick's, a U.S.\ supermarket chain. 
	Sales, price, and promotion data are collected from January 1993 to July 1994, hence $T=76$, and are publicly available from \url{https://research.chicagobooth.edu/kilts/marketing-databases/dominicks}. 
	For more information on the calculation of the sales, price, and promotion variables, see  \cite{Srinivasan04}.
	We apply the HVARX model with $k=16$ endogenous sales series and $m=32$ exogenous price and promotion series, with $p=s=\lfloor 1.5\sqrt{T}\rfloor=13$.
	
	\paragraph{Estimated Demand Effects.}
	Since HVARX performs automatic lag selection, it returns the effective maximum lag orders of each elementwise component. 
	Consider the $k\times k$ endogenous lag matrix $\widehat{L}_{\widehat{\Phi}}$ and $k\times m$ exogenous lag matrix $\widehat{L}_{\widehat{B}}$ of an estimated VARX with elements 
	$$\widehat{L}_{\widehat{\Phi}, id} = \max\{\ell: \widehat{\Phi}_{\ell, id} \neq 0\}, \  \ \ \text{and} \  \ \  \widehat{L}_{\widehat{B}, ir} = \max\{s: \widehat{B}_{s, ir} \neq 0\}, $$
	and  $\widehat{L}_{\widehat{\Phi}, id}=0$ if $\widehat{\Phi}_{\ell, id}=0$ for all $\ell=1,\ldots,p$; analogously for $\widehat{L}_{\widehat{B}}$.  
	The lag matrix $\widehat{L}_{\widehat{\Phi}}$ gives the maximal lag for each endogenous series $d$ in each marginal equation $i$ for an estimated VARX. Similarly, $\widehat{L}_{\widehat{B}}$
	gives the maximal lag for each exogenous series $r$ in each equation $i$.
	
	\begin{figure}
		\includegraphics[width=14cm]{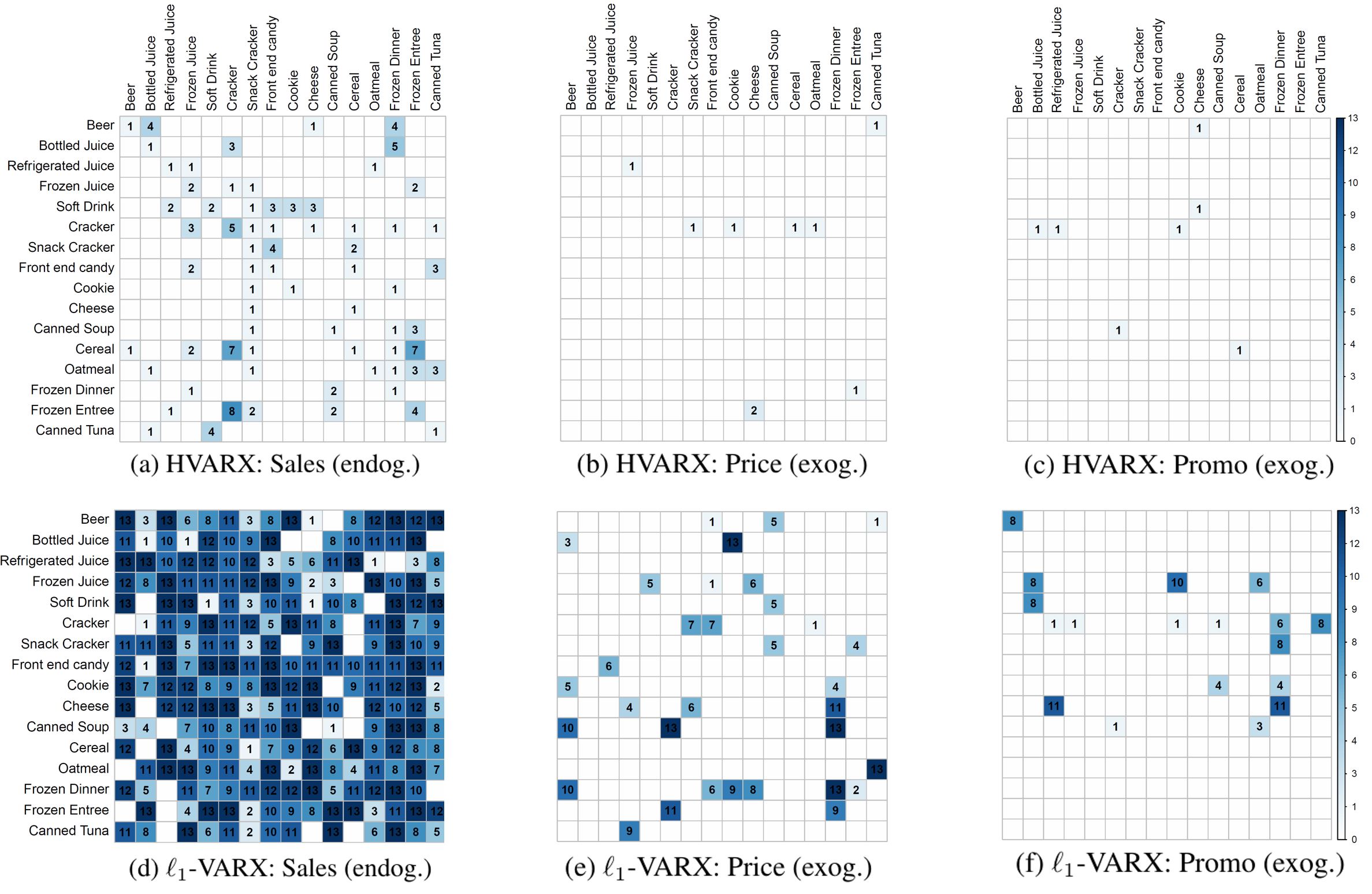}
		\caption{Panel (a) and (d): Lag matrix of endogenous sales time series, panel (b) and (e): exogenous price time series and panel (c) and (f):  exogenous promotion time series, for the HVARX (top) and the $\ell_1$-VARX (bottom) estimators, respectively. The maximum orders in the VARX were $\lfloor 1.5\sqrt{T} \rfloor =13$.\label{Lhats}}
	\vspace{-6pt}
	\end{figure}

	Figure \ref{Lhats} compares the estimated lag matrices of HVARX (top) to those of  $\ell_1$-VARX (bottom).
	The first column gives the lag matrices of the endogenous sales time series (i.e., $\widehat{L}_{\widehat{\Phi}}$), 
	the second column gives the lag matrices of the exogenous price time series, and 
	the third column gives the lag matrices of the exogenous promotion time series. 
	Note that panels (b)-(c); (e)-(f) combined give the $16\times32$-matrix $\widehat{L}_{\widehat{B}}$.
	The lag matrices of HVARX are overall sparser than those of $\ell_1$-VARX, with considerably lower maximal lag orders for the endogenous sales time series. The Bayesian Information Criterion confirms the model estimated by HVARX to provide a better in-sample fit  than the model estimated by $\ell_1$-VARX: BIC=11.08 versus BIC=43.11, respectively.
	
	By encouraging the maximal lag orders to be small, HVARX gives a highly interpretable model. The lag matrix of the endogenous time series (panel a) shows the prevalence of within-category sales effects: each sales product category (except cheese) is  influenced by its own past sales values (i.e., non-zero diagonal elements). Only a minority of cross-category sales effects  are estimated as non-zero (i.e., non-zero off-diagonal elements). 
	The lag matrices of exogenous time series (panel b and c) are very sparse: for HVARX,
	8 out of 256 entries of the price lag matrix and 7 out of 256 entries of the promotion lag matrix are non-zero.  
	These lag matrices mainly reveal cross-category effects possibly due to substitution or affinity in consumption.
	The more parsimonious model obtained with HVARX also gives more accurate forecasts, as discussed in the following paragraph.
	
	\paragraph{Forecast Accuracy.}
	We compare the forecast accuracy of HVARX  and $\ell_1$-VARX. 
	To assess forecast accuracy, we use an expanding window forecast approach (see \citealp{Wilms17}) where we compute one-step-ahead forecasts for the last 15\% of the observations. 
	The overall forecast performance is then measured by the Mean Squared Forecast Error  (averaged over all time series and all time points). 
	HVARX attains a lower MSFE than $\ell_1$-VARX: MSFE=0.391 versus MSFE=0.453. 
	A Diebold-Martiano \citep{Diebold95} test also confirms this improvement in forecast accuracy to be significant at the 5\% significance level.
		
	\subsubsection*{Acknowledgments} IW was supported by the Research Foundation-Flanders (FWO-12M8217N),
	JB was supported by an NSF CAREER award (DMS-1748166),
	DSM was supported by an NSF CAREER award (DMS-1455172), a Xerox PARC Faculty Research Award, and Cornell University Atkinson Center for a Sustainable Future (AVF-2017).

	\begin{small}
		\bibliographystyle{asa}
		\bibliography{VARMAref}
	\end{small}

\end{document}